\documentclass{llncs}
\usepackage{xstring}
\usepackage[T1]{fontenc}
\usepackage{graphicx}
\usepackage{amsmath}
\usepackage{amsfonts}
\usepackage{subcaption}
\usepackage{placeins}
\usepackage{float}
\usepackage{tikz}
\usetikzlibrary{arrows.meta, positioning}
\usepackage{algorithm}
\usepackage{enumitem}
\usepackage{comment}
\usepackage{booktabs}
\usepackage[noend]{algpseudocode}
\usepackage{url}
\usepackage{hyperref}  
\usepackage{orcidlink}  

\begin{document}

\title{Genetic Generalized Additive Models}
\titlerunning{\textbf{Genetic Generalized Additive Models}}

\author{
Kaaustaaub Shankar\inst{1}\orcidlink{0009-0000-8970-3746} \and
Kelly Cohen\inst{1}\orcidlink{0000-0002-8655-1465}
}

\institute{
College of Engineering and Applied Science, University of Cincinnati, Cincinnati, OH 45219, USA\\
\email{shankaks@mail.uc.edu}
}

\authorrunning{K.Shankar et al.}

\maketitle
\newcommand{\featurefigure}[2]{%
  \begin{figure}[htbp]
    \centering
    \includegraphics[width=0.85\linewidth]{#1}
    \caption{#2}
    \label{fig:#1}
  \end{figure}%
}

\abstract{Generalized Additive Models (GAMs) balance predictive accuracy and interpretability, but manually configuring their structure is challenging. We propose using the multi-objective genetic algorithm NSGA-II to automatically optimize GAMs, jointly minimizing prediction error (RMSE) and a Complexity Penalty that captures sparsity, smoothness, and uncertainty. Experiments on the California Housing dataset show that NSGA-II discovers GAMs that outperform baseline LinearGAMs in accuracy or match performance with substantially lower complexity. The resulting models are simpler, smoother, and exhibit narrower confidence intervals, enhancing interpretability. This framework provides a general approach for automated optimization of transparent, high-performing models. The code can be found at \url{https://github.com/KaaustaaubShankar/GeneticAdditiveModels}.
}

\FloatBarrier
\section{Introduction}
As machine learning systems increasingly find their way into high-stakes domains, from housing valuation to autonomous weapons, the need for models that are not just accurate but inherently interpretable has never been more urgent. Researchers such as Missy Cummings have argued that AI systems should never be entrusted with control over life-and-death decisions unless their decision-making logic can be tested for wrong outputs\cite{cummings2025prohibiting} and interpretable models can help resolve this. This perspective regarding safety and in general how we integrate machine learning into society has been explored by many like Cynthia Rudin who believes that the best models are models that are inherently interpretable rather than explained by post-hoc methods that are unreliable since they are post-hoc\cite{rudin2019stopexplainingblackbox}.

As regulations around the world (an example being the EU AI Act\cite{EU2024AIActRecital72}) urge AI developers and services built on these models to think more about how to make their product safer, the demand for transparent models has expanded far beyond defense and encompasses much of our day-to-day life. However, designing these systems is a significant challenge; their structure and hyperparameters have to be specified ahead of time and these choices determine their approximation capabilities, how well they generalize, and how interpretable they are. This is true for models like generalized additive models\cite{gam} and fuzzy inference systems (Takagi-Sugeno-Kang\cite{TSK} and Mamdani\cite{MAMDANI19751}) where the structure of these models is the structure of the explanation. Selecting an appropriate structure is therefore not just about finding the best structure to reduce error but a principled way of controlling the complexity, inductive biases, and human comprehensibility of the learned representation. 

Furthermore, to keep models human-interpretable, we need to introduce sparsity constraints that limit the complexity of the model. By controlling which components are active and how flexible each component can be, we reduce cognitive load and make the model easier to understand and audit. In addition, sparsity helps ensure that each part of the model contributes meaningful information, improving both the clarity and reliability of the explanations without relying solely on accuracy metrics.

Hence, this work explores the use of a genetic algorithm, specifically NSGA-II\cite{NSGAII}, to automatically optimize the structural hyperparameters of GAMs, enabling models that are both accurate and meaningfully interpretable.
\FloatBarrier
\section{Related Work}
\subsection{Generalized Additive Models (GAMs)}
Generalized Additive Models (GAMs)\cite{gam} are a class of interpretable models that decompose a prediction into a sum of univariate or independent effects. This can be represented as the following:
$$
y = \beta_0 + \sum_{j=1}^p f_j(x_j)
$$
where $f_j$ is a smooth function applied to a single feature. Through this assumption, GAMs are highly interpretable since we can visualize the contribution of each feature independently. However, defining $f_j$ remains complicated, as real-world relationships may require flexible or nonlinear functions. To accommodate this, each $f_j$ can be specified as either a simple linear term or as a spline-based function.

Linear terms provide fully transparent, globally monotonic behavior but lack the expressiveness needed when the data exhibits nonlinear structure. Spline terms, in contrast, allow $f_j$ to adapt smoothly to nonlinear relationships while maintaining interpretability. These spline functions are built from a set of basis functions, and their flexibility is controlled by two key hyperparameters: the number of spline basis functions and the regularization constant $\lambda$.

The number of basis functions determines how much curvature the model can express. The regularization constant $\lambda$ further controls smoothness: a large $\lambda$ forces $f_j$ to be nearly linear, whereas a small $\lambda$ allows for high variability. These hyperparameters therefore play a structural role, directly influencing the expressiveness and interpretability of the resulting model.

Despite their simple nature, they have proven to be effective and a competitive choice against traditional models like neural networks for tabular data in a variety of fields \cite{doohan2025comparisongeneralisedadditivemodels}. 

\subsection{Genetic Algorithms}
Genetic algorithms \cite{holland1975adaptation} are a class of evolutionary algorithms that optimize a population of candidate solutions using biologically inspired operators such as crossover and mutation. By evolving an entire population rather than a single point estimate, GAs perform a broad exploration of the search space and help avoid the local minima that often trap greedy or gradient-based methods. This makes them particularly effective for multi-objective optimization, where different objectives may conflict. Approaches such as CMA-ES \cite{CMAES} and NSGA-II \cite{NSGAII} demonstrate how evolutionary methods can jointly optimize multiple criteria and approximate a Pareto front which is the set of solutions that are non-dominated with respect to all objectives. In the context of interpretable modeling, this provides a natural mechanism for balancing predictive accuracy against model simplicity: the GA can search over a structured model space and return a spectrum of trade-offs rather than a single solution. From this spectrum, we can identify models that best align with regulatory requirements such as models that maintain strong predictive accuracy while ensuring structural simplicity and interpretability.

\subsection{Pareto Front}
A central concept in multi-objective optimization is the \emph{Pareto front}, the set of solutions for which no objective can be improved without degrading another. Instead of producing a single optimal model, evolutionary algorithms such as NSGA-II identify a frontier of non-dominated solutions that represent different trade-offs between competing goals. Prior work has shown that Pareto-based search is effective in balancing predictive accuracy with sparsity, interpretability, or computational cost, making it well suited for model-selection settings where multiple criteria must be jointly optimized\cite{schneider2023multiobjectiveoptimizationperformanceinterpretability}\cite{Blanzeisky_Cunningham_2022}.

\section{Methodology}
We employ NSGA-II to search over GAM structures and hyperparameters. The GA will evolve these models using mutation and crossover, and the models are evaluated using predictive performance (RMSE) and a complexity penalty promoting sparsity and interpretability.

These models are trained on the California Housing dataset and are compared to baseline LinearGAMs and Decision Trees over multiple seeds. 

\subsection{NSGA-II}

Each candidate model is encoded as a chromosome, where genes represent the type of term for each feature (none, linear, or spline), the number of splines, the smoothing parameter $\lambda$, and whether to scale the feature. Chromosomes were initialized by the following:
\begin{algorithm}[htbp]
\caption{Smart Initialization of GAM Chromosome}
\begin{algorithmic}[1]
\State \textbf{Input:} Number of features $n$, maximum splines $K_{\text{max}} = 20$
\State \textbf{Output:} Chromosome $\text{genes}$
\State $\text{genes} \gets []$
\For{$i = 1$ to $n$}
    \State Sample component type $\text{type} \in \{\text{none, linear, spline}\}$ with probabilities $[0.2, 0.3, 0.5]$
    \If{$\text{type} = \text{spline}$}
        \State $\text{knots} \gets \text{random integer in } [8, K_{\text{max}}]$
        \State $\lambda \gets \text{random float in } [0.1, 10.0]$
    \Else
        \State $\text{knots} \gets \text{None}$, $\lambda \gets \text{None}$
    \EndIf
    \State $\text{scale} \gets \text{random choice } \{\text{True, False}\}$
    \State Append $\{\text{type, knots, }\lambda, \text{scale}\}$ to $\text{genes}$
\EndFor
\State \Return $\text{genes}$
\end{algorithmic}
\end{algorithm}

Uniform crossover and perturbation mutation were employed in the genetic algorithm. Crossover was applied with a probability of 0.3, while mutation used an adaptive rate starting at 0.15 and gradually decreasing over 100 generations. The population size was set to 80 individuals, and the algorithm was run for 50 generations. These values were selected based on preliminary tuning to balance convergence quality with computational cost.

NSGA-II is tasked with optimizing two objectives: RMSE and model complexity 
$C \in [0,1]$, defined as a weighted combination of parameter uncertainty and model sparsity:
\begin{equation}
C = 0.70\,U + 0.30\,S.
\end{equation}
The uncertainty score $U$ is the mean width of the 95\% confidence intervals of all active model terms, normalized by the range of a reference target $y_{\text{ref}}$,
\begin{equation}
U = \frac{1}{n_{\text{active}}} \sum_{j=1}^{n_{\text{active}}}
\frac{\text{CIwidth}_j}{\max(\text{range}(y_{\text{ref}}), \epsilon)},
\end{equation}
where $n_{\text{active}}$ denotes the number of active terms and $\epsilon$ avoids division by zero; terms without valid confidence intervals are assigned a normalized width of 1. 
The sparsity fraction $S$ is defined as the proportion of active terms relative to the total number of available features,
\begin{equation}
S = \frac{\min(n_{\text{active}}, n_{\text{features}})}{n_{\text{features}}}.
\end{equation}

By minimizing the sparsity function as well as the uncertainty score, a less complex but more interpretable model can be found.

For each individual, we perform cross-validation on the combined training and validation data to obtain a robust estimate of performance. The final error is averaged across all cross-validation folds (k = 5). 

To capture the range of achievable trade-offs, we extract three representative Pareto-optimal models: the knee solution, the RMSE-optimal model, and the model with the minimal complexity penalty.

\subsection{California Housing Dataset}

The California Housing dataset is a standard benchmark for regression tasks. It contains information collected from the 1990 U.S. Census, aggregated by block group. The dataset consists of 8 features and 1 continuous target variable (median house value in hundreds of thousands of dollars). Table~\ref{tab:california_features} summarizes the features.

\begin{table}[htbp]
\centering
\begin{tabular}{ll}
\hline
\textbf{Feature} & \textbf{Description} \\
\hline
MedInc & Median income in block group (in tens of thousands of dollars) \\
HouseAge & Median age of houses in the block group \\
AveRooms & Average number of rooms per household \\
AveBedrms & Average number of bedrooms per household \\
Population & Total population of the block group \\
AveOccup & Average number of occupants per household \\
Latitude & Block group latitude \\
Longitude & Block group longitude \\
\hline
\textbf{Target} & Median house value (in hundreds of thousands of dollars) \\
\hline
\end{tabular}
\caption{Features and target of the California Housing dataset.}
\label{tab:california_features}
\end{table}
\FloatBarrier
\subsection{Baseline Models}
The results of our GA-trained models will be compared to a LinearGAM where all features will be using spline functions and will have a max of 25 splines each whereas the decision tree will have no maximum depth. The decision tree acts as an upper bound, high-variance baseline.
\FloatBarrier
\section{Results}

\begin{table}[htbp]
\centering
\setlength{\tabcolsep}{6pt} 
\begin{tabular}{r|ccccc}
\hline
\textbf{Seed} & \textbf{GAM (rmse)} & \textbf{GAM (knee)} & \textbf{GAM (penalty)} & \textbf{Decision Tree} & \textbf{LinearGAM} \\
\hline
42 & \textbf{0.6821} & 0.7125 & 0.7400 & 0.6925 & 0.7216 \\
7 & \textbf{0.6554} & 0.6745 & 0.7246 & 0.7159 & 0.6674 \\
123 & \textbf{0.6523} & 0.6663 & 0.7130 & 0.6961 & 0.6697 \\
225 & \textbf{0.6228} & 0.6702 & 0.7824 & 0.6770 & 0.6536 \\
729 & \textbf{0.6496} & 0.6890 & 0.7108 & 0.6801 & 0.6520 \\
\hline
\end{tabular}
\caption{Test RMSE scores for five different models across various random seeds. The GA models are optimized for: lowest RMSE, the "knee" (best trade-off), and lowest complexity penalty. Bold values indicate the overall best performing model for each seed.}
\label{tab:rmse_results}
\end{table}

\begin{table}[htbp]
\centering
\setlength{\tabcolsep}{4pt} 
\begin{tabular}{r|cccc}
\hline
\textbf{Seed} & \textbf{Baseline} & \textbf{GAM (knee)} & \textbf{GAM (best\_by\_rmse)} & \textbf{GAM (best\_by\_penalty)} \\
\textbf{} & \textbf{Penalty} & \textbf{Penalty} & \textbf{Penalty} & \textbf{Penalty} \\
\hline
42 & 0.4096 & 0.1020 & 0.2226 & \textbf{0.0495} \\
7 & 0.4384 & 0.1434 & 0.2390 & \textbf{0.0512} \\
123 & 0.4562 & 0.1450 & 0.3986 & \textbf{0.0497} \\
225 & 0.4353 & 0.1083 & 0.2905 & \textbf{0.0549} \\
729 & 0.4608 & 0.0689 & 0.2508 & \textbf{0.0491} \\
\hline
\end{tabular}
\caption{Complexity penalty scores for different models across various random seeds. Bold values indicate the lowest penalty score for each seed.}
\label{tab:penalty_results_narrow}
\end{table}

\begin{figure}[htbp]
    \centering
    \includegraphics[width=0.7\textwidth]{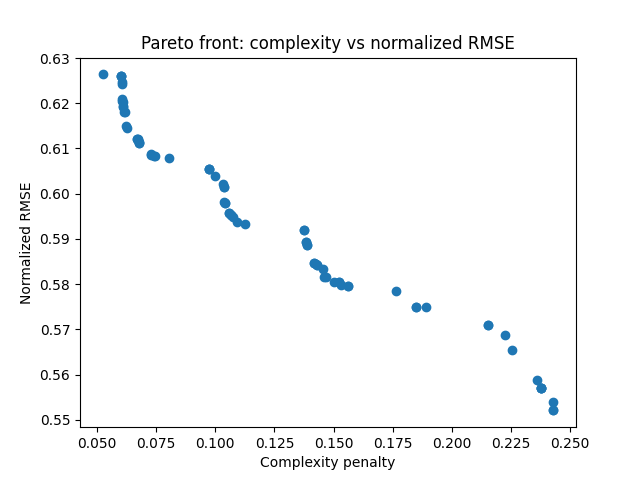}
    \caption{Pareto Front for seed 7}
    \label{fig:improved_pareto_front_seed7}
\end{figure}

\begin{figure}[htbp]
    \centering
    \includegraphics[width=0.7\textwidth]{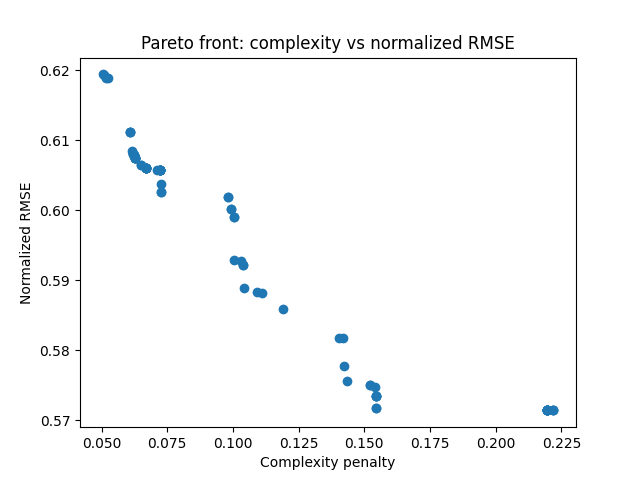}
    \caption{Pareto Front for seed 42}
    \label{fig:improved_pareto_front_seed42}
\end{figure}

\begin{figure}[htbp]
    \centering
    \includegraphics[width=0.7\textwidth]{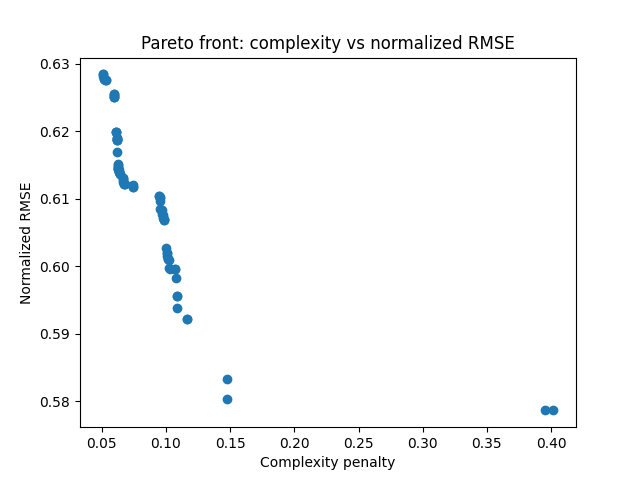}
    \caption{Pareto Front for seed 123}
    \label{fig:improved_pareto_front_seed123}
\end{figure}

\begin{figure}[htbp]
    \centering
    \includegraphics[width=0.7\textwidth]{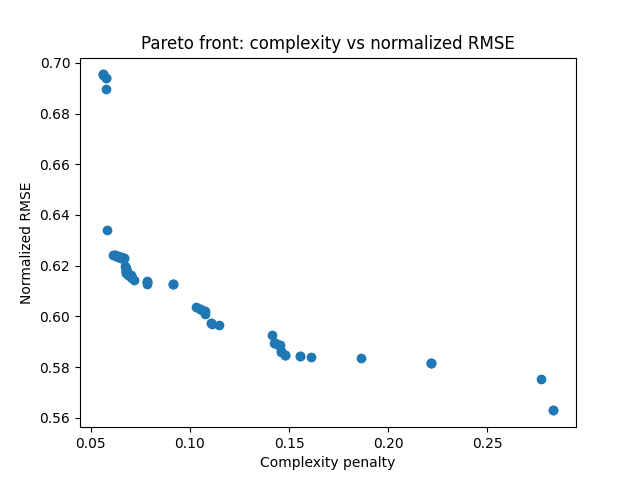}
    \caption{Pareto Front for seed 225}
    \label{fig:improved_pareto_front_seed225}
\end{figure}

\begin{figure}[htbp]
    \centering
    \includegraphics[width=0.7\textwidth]{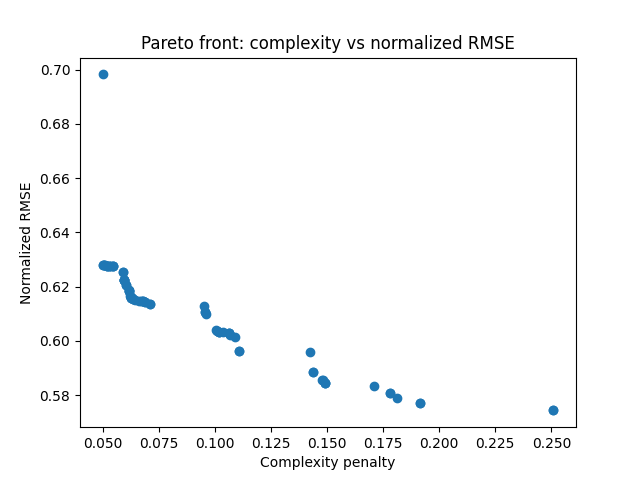}
    \caption{Pareto Front for seed 729}
    \label{fig:improved_pareto_front_seed729}
\end{figure}

\begin{figure}
    \centering
    \includegraphics[width=0.9\linewidth]{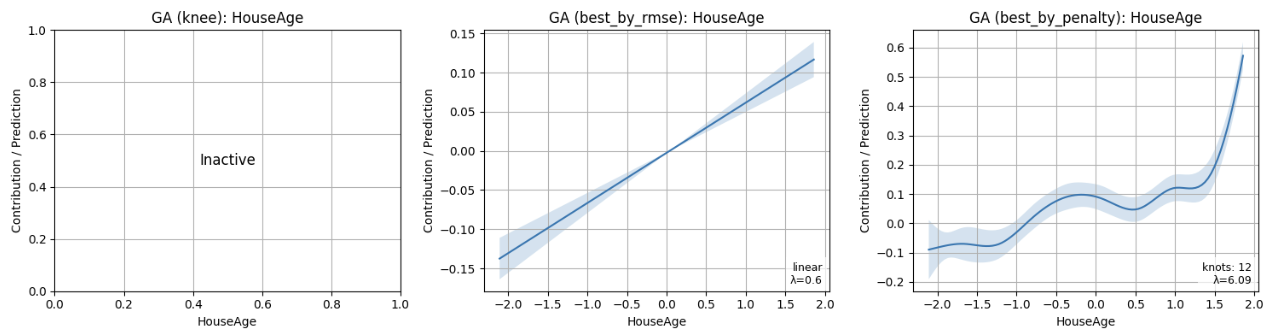}
    \caption{GA GAM Feature Contributions for House Age for Seed 7}
    \label{fig:gagam_seed7_houseage}
\end{figure}
\begin{figure}
    \centering
    \includegraphics[width=0.7\linewidth]{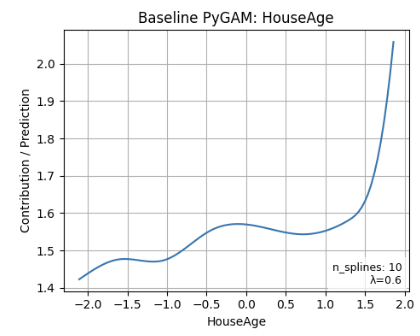}
    \caption{Linear Gam Feature Contributions for House Age for Seed 7}
    \label{fig:lineargam_seed7_houseage}
\end{figure}
\FloatBarrier
\section{Discussion}
The results in Table \ref{tab:rmse_results} illustrate the characteristic behavior of a two-objective NSGA-II optimization applied to GAM structure search. As expected, the GA model selected for best RSME consistently achieves the lowest test error across all seeds, outperforming both the Decision Tree and the baseline LinearGAM. This reflects the RMSE-dominant region of the Pareto front, where the algorithm prioritizes predictive accuracy at the expense of increased complexity and uncertainty. In contrast, the GA model selected for minimal penalty exhibits the highest RMSE among the GA solutions yet remains competitive against the baselines. The knee-point model, which represents the best global trade-off between prediction error and the penalty term, consistently occupies an intermediate position: it is less accurate than the RMSE-optimal model but typically comparable to or better than the baseline LinearGAM while maintaining substantially lower complexity. These results showcase how NSGA is able to find solutions for both of these objectives.

In conjunction with Table \ref{tab:penalty_results_narrow}, the results demonstrate that NSGA-II is able to discover models that achieve competitive predictive performance while maintaining substantially lower complexity. Notably, the GA model picked for the lowest RMSE is consistently simpler than the baseline LinearGAM, despite outperforming it in prediction accuracy across all seeds. This indicates that within the space of high-performing GAMs, NSGA-II effectively identifies solutions that reside on a favorable portion of the Pareto front—models that provide strong predictive accuracy without incurring the unnecessary structural complexity typically produced by fully unconstrained spline fits. Consequently, the results reinforce the idea that meaningful interpretability gains can be achieved without sacrificing predictive performance, provided that the model search is guided by a multi-objective framework like NSGA-II.

Figures \text{\ref{fig:gagam_seed7_houseage}}, \text{\ref{fig:lineargam_seed7_houseage}} display how the GAM when guided by a multi-objective optimization achieves a model that is both structurally simpler and more honest about its predictive certainty compared to a standard baseline. For features like HouseAge (as referenced), the NSGA-GAM achieves the ultimate interpretability gain: sparsity. By setting the effect to "Inactive," the model removes the need to interpret the complex, wiggly spline chosen by the Baseline PyGAM, which might have been chasing noise. Crucially, in features where the NSGA model does simplify the function to linear, the resulting confidence interval (CI) width is highly informative; the intervals often flare out significantly at the extremes. This flaring is a feature, not a bug, as it signals an honest uncertainty to the user in sparse, extrapolated data regions, thus preventing the false precision inherent in the Baseline's tight, locally-fitted CIs, which is a safer outcome for high-stakes decision-making.
Finally, the Pareto fronts for all seeds (Figures \ref{fig:improved_pareto_front_seed7}, \ref{fig:improved_pareto_front_seed42}, \ref{fig:improved_pareto_front_seed123}, \ref{fig:improved_pareto_front_seed225}, \ref{fig:improved_pareto_front_seed729}) illustrate that each run yields a diverse set of high-quality solutions. Rather than identifying a single “best’’ model, NSGA-II produces a spectrum of models that balance accuracy and complexity in different ways. This is particularly important in practice, where the preferred solution often depends on external requirements such as interpretability, stability, or regulatory constraints. The results emphasize that model selection cannot rely on accuracy alone; instead, multiple objectives must be jointly considered to ensure that the chosen model is not only predictive but also appropriate for the operational and regulatory context.

\FloatBarrier
\section{Future Work}
This work shows that NSGA-II can successfully optimize GAM structures by balancing accuracy with interpretability-oriented objectives. A natural next step is to extend this multi-objective approach to fuzzy and cascading rule-based systems, which pursue similar goals but involve different design choices. Fuzzy systems depend on decisions such as membership function shapes, the number and structure of rules, and the layout of cascading stages. Each of these can be treated as a structural hyperparameter, similar to knots or smoothing levels in GAMs. Future research could explore how the same types of objectives used here, such as accuracy, confidence interval width, sparsity, and rule simplicity, can guide the automated design of fuzzy models using evolutionary search. Applying NSGA-II to the design of layered or cascading fuzzy architectures may help create systems that remain interpretable while adapting their structure to the data in a principled way.

Another practical direction is to reduce the computational cost of evaluating candidate models. Fitting a GAM for every individual in every generation can be expensive, and this cost grows as we explore richer model spaces. Future work can introduce caching or memoization strategies that store and reuse the results of previously evaluated structures. This would avoid repeated model fits for identical or near-identical chromosomes and would make large-scale multi-objective optimization more feasible. Together, these extensions would broaden the impact of NSGA-II for interpretable model design across multiple modeling paradigms.
\FloatBarrier

\FloatBarrier
\bibliographystyle{unsrt}
\bibliography{Bib}

\FloatBarrier
\end{document}